# SINGULAR VALUE DECOMPOSITION OF IMAGES FROM SCANNED PHOTOGRAPHIC PLATES

VASIL KOLEV[1], KATYA TSVETKOVA[2] and MILCHO TSVETKOV[3]

[1]*Institute of Information and Communications Technology*
*Bulgarian Academy of Sciences*
E-mail: kolev_acad@abv.bg
[2]*Institute of Mathematics and Informatics*
*Bulgarian Academy of Sciences*
E-mail: katya@skyarchive.org
[3]*Institute of Astronomy with National Astronomical Observatory*
*Bulgarian Academy of Sciences*
E-mail: milcho@skyarchive.org, ana@skyarchive.org

**Abstract.** We want to approximate the $m \times n$ image **A** from scanned astronomical photographic plates (from the Sofia Sky Archive Data Center) by using far fewer entries than in the original matrix. By using rank of a matrix, $k$ we remove the redundant information or noise and use as Wiener filter, when rank $k < m$ or $k < n$. With this approximation more than 98% compression ration of image of astronomical plate without that image details, is obtained. The SVD of images from scanned photographic plates (SPP) is considered and its possible image compression.

## 1. INTRODUCTION

The need to minimize the amount of digital information stored and retrieved is an ever growing concern in the modern world. Singular Value Decomposition (SVD) (Andrews and Patterson, 1976) is an effective tool for minimizing data storage and data transfer. Application of SVD in astronomy can be found in (Boissel et al., 2001), where SVD is applied to a mid-infrared ISOCAM spectral map of NGC 7023 and a mathematical analysis of the map in terms of a linear combination of elementary spectra is provided. The spectrum observed on each pixel can be described as the physical superposition of four components - the intrinsic spectra of polycyclic aromatic hydrocarbons, very small grains, larger dust grains and a differential spectrum that could trace the ionisation state of polycyclic aromatic hydrocarbons.

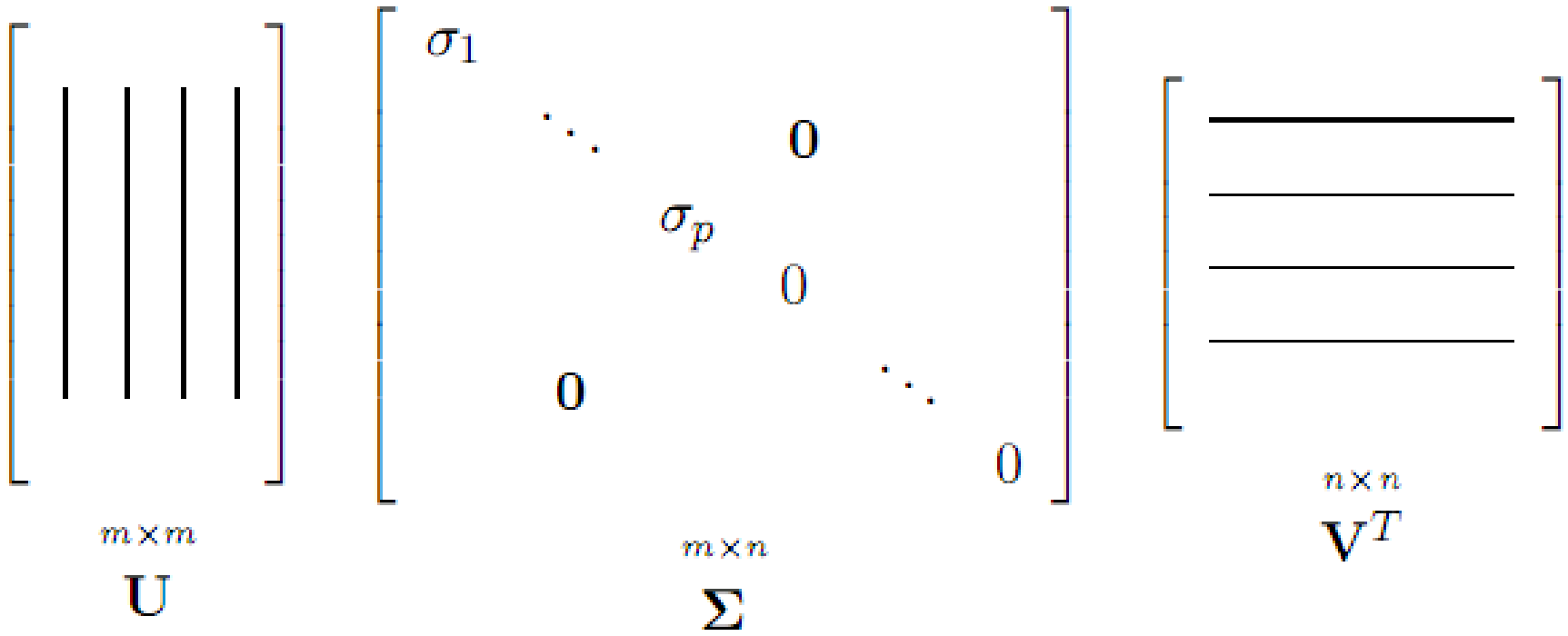

**Figure 1.** Structure of SVD matrices decomposition.

Other application of SVD is made for separation of image data and noise subspaces using SVD (Yatawatta, 2008). The SVD characterized the signal and noise subspace eigenmodes. Because the noise has much lower power compared with the signal, the eigenmodes corresponding to the dominant singular values.

SVD is applied also for detection of faint stars, noise removing, continuum subtraction of spectral lines for radio-astronomical images, and automatic image classification.

The goal of this paper is an application of SVD as a new approach for images analysis from scanned photographic plates (SPPs). This approach is in connection with a future creation of the image compression database of Rozhen Observatory SPPs.

The SPPs stored on the servers of the Sofia Sky Archive Data Center, are with large image sizes, which take a lot of space in the computer systems. In order to minimizing such image sizes there are different methods for image compression. One such method is a Singular Value Decomposition - very useful technique in data analysis and visualization. In linear algebra SVD is a well-known technique for factorizing a rectangular matrix, real or complex, which has been widely employed in signal processing, like image compression (Demmel, 1997; Nievergelt, 1997), noise reduction or image watermarking.

Recently, the SVD transform was used to measure the image quality under different types of distortions (Shnayderman, Gusev and Eskicioglu, 2004). Among all useful decompositions SVD - that is the factorization of a matrix **A** into the product $\mathbf{U\Sigma V}^\mathbf{T}$ of a unitary matrix **U**, diagonal matrix $\mathbf{\Sigma}$, and another matrix $\mathbf{V}^\mathbf{T}$ - assumes a special role (Fig.1). There are several reasons for it:

- The fact that the decomposition is achieved by unitary matrix, makes it an ideal vehicle for discussing the geometry of n-space;

- SVD is stable, small perturbation in **A** correspondent to small perturbation in $\mathbf{\Sigma}$ and conversely;

- The diagonality of $\mathbf{\Sigma}$ makes it easy to determine when **A** is near to rank-degenerate matrix, and when it is, the decomposition provides optimal low-rank approximation to **A;**

- Thanks to the pioneering efforts of Gene Golub, efficient and stable algorithms to compute the SVD have already existed.

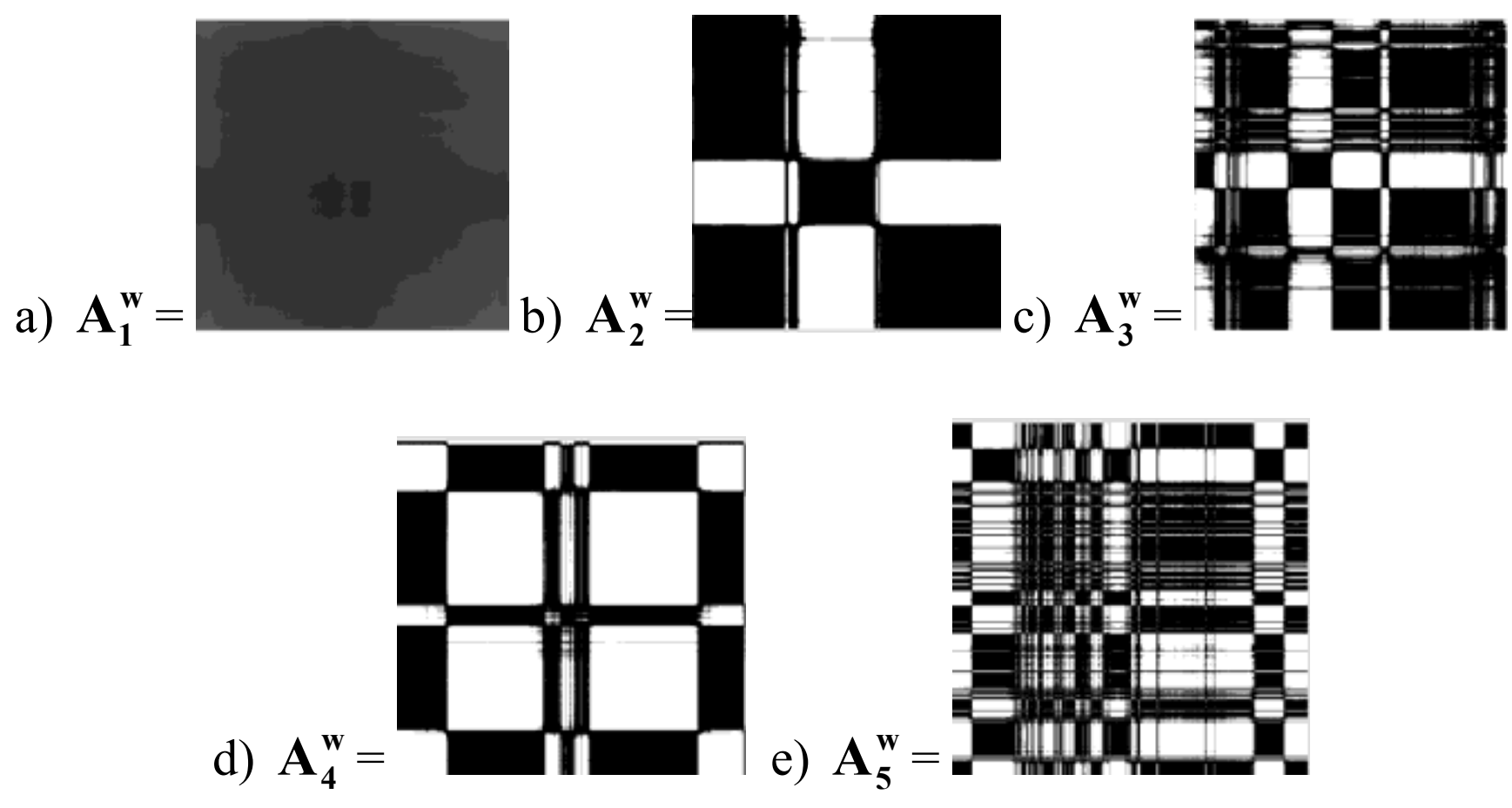

**Figure 2.** A weight matrix $\mathbf{A}_j^w = u_j v_j^T$, $j = 1,2..5$

$$= 794.73 * \mathbf{A}_1^w + 71.841 * \mathbf{A}_2^w + 30.338 * \mathbf{A}_3^w + 29.161 * \mathbf{A}_4^w + 16.107 * \mathbf{A}_5^w$$

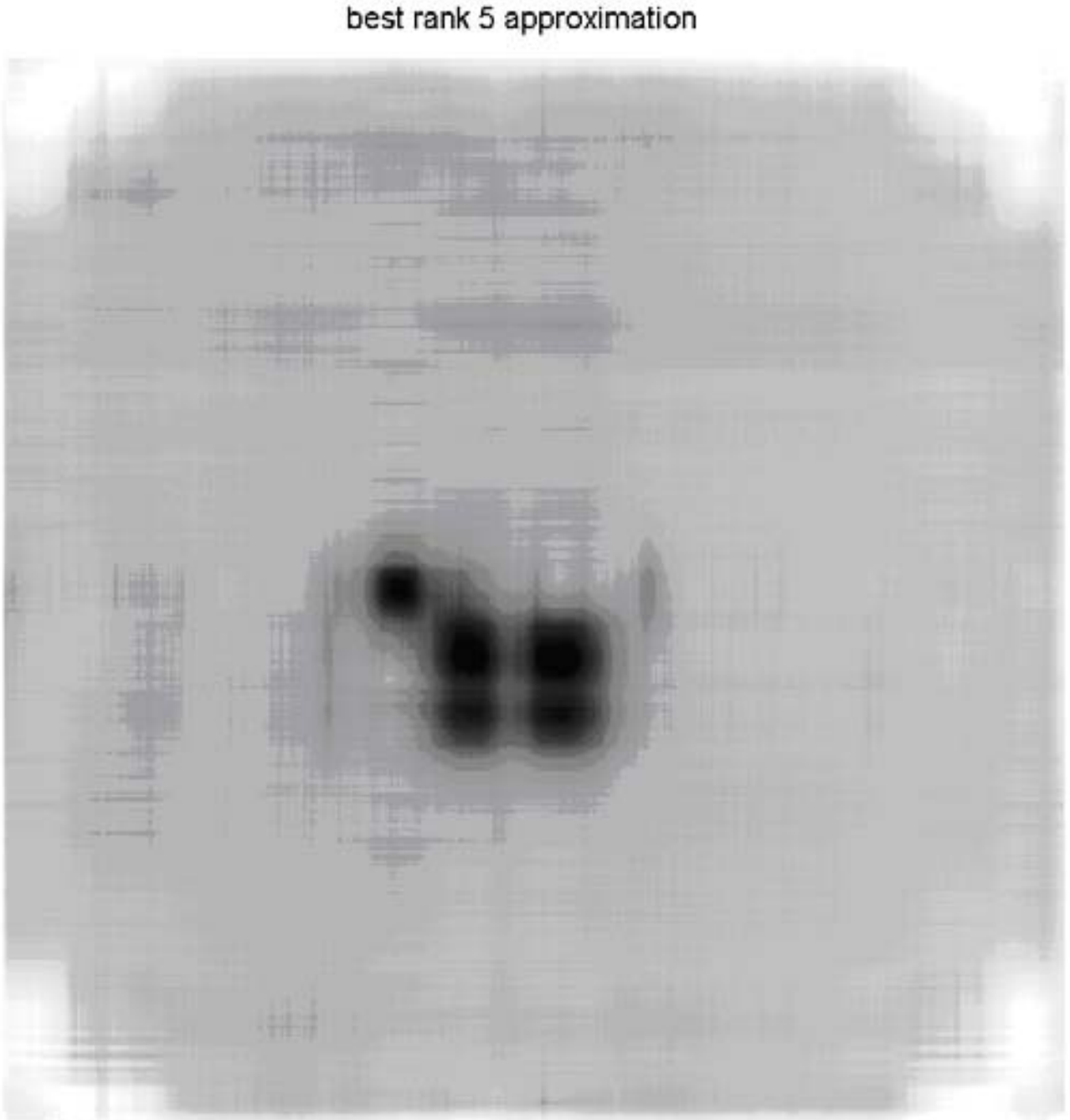

**Figure 3.** A weight matrix image decomposition of SPP ASI067 000556 (M45-556p.fits) in the region of the Pleiades stellar cluster.

SVD is an intriguing analogy between reduced rank approximations and Fourier analysis. Particularly in the discrete case Fourier analysis can be viewed as representing a data vector relative to a special orthogonal basis. The basis elements are envisioned as pure vibrations, that is sine and cosine functions, at different frequencies. The Fourier decomposition thus represents the input data as a superposition of pure vibrations with the coefficients specifying the amplitude of each constituent frequency. Often, there are a few principal frequencies that account for most of the variability in the original data. The remaining frequencies can be discarded with little effect. The reduced rank approximations based on the SVD are very similar in intent. However, SVD captures the best possible basis vectors for the particular data observed, rather than using one standard basis for all cases. For this reason, SVD - based reduced rank approximation - can be thought of as an adaptive generalization of Fourier analysis. The most significant vibrations are adapted to the particular data that appear.

## 2. SINGULAR VALUES AND THE MATRIX 2-NORM

Let us introduce matrix 2-norm for real –value matrix $\mathbf{A}$ :

$$\|\mathbf{A}\|_2 = \max_{\|x\|_2=1} \|\mathbf{A}x\|_2 = \sqrt{\lambda_{max}} \tag{1}$$

where $x$ is a vector and $\lambda_{max}$ is the largest eigenvalue such that $\mathbf{A}^\mathbf{T}\mathbf{A} - \lambda\mathbf{I}$ is singular.

The matrix 2-norm inherits unitary invariance from the vector 2-norm: for any unitary matrices $\mathbf{U}$ and $\mathbf{V}$, $\|\mathbf{UAV}\|_2 = \|\mathbf{A}\|_2$, but did not provide a simple formula for this norm in terms of the entries of A, as we did for the induced matrix 1- and $\infty$-norms. With the SVD we can now derive such a formula. Recall that the vector 2-norm (and hence the matrix 2-norm) is invariant to premultiplication by a unitary matrix. Let $\mathbf{A} = \mathbf{U\Sigma V^T}$ be a singular value decomposition of A.

This $\|\mathbf{A}\|_\mathbf{2} = \|\mathbf{U\Sigma V^T}\|_\mathbf{2} = \|\mathbf{\Sigma V^T}\|_\mathbf{2}$. The matrix 2-norm is also immune to a unitary matrix on the right:

$$\|\mathbf{\Sigma V^T}\|_\mathbf{2} = \max_{\|x\|_2=1} \|\mathbf{\Sigma V^T} x\|_\mathbf{2} = \max_{\|y\|_2=1} \|\mathbf{\Sigma} y\|_\mathbf{2} = \|\mathbf{\Sigma}\|_\mathbf{2}, \tag{2}$$

where we have set $y = \mathbf{V^T} x$ and noted that $\|y\|_2 = \|\mathbf{V^T} x\|_2 = \|x\|_2$, since $\mathbf{V^T}$ is unitary matrix.

Let $p = \min\{m, n\}$, where m-size matrix $\mathbf{U}$, n-size matrix $\mathbf{V}$, Fig.1. Then $\|\mathbf{\Sigma}y\|_2^2 = \sum_{j=1}^{p} \sigma_j^2 y_j^2$ which is maximized over $\|y\|_2 = 1$ by $y = [1,0,...,0]^T$, giving

$\|\mathbf{A}\|_2 = \|\mathbf{\Sigma}\|_2 = \sigma_1$. Thus the matrix 2-norm is simply the first singular value and it is the largest singular value. Examples of singular values of full rank are shown in Fig. 4-6 and without zero eigenvalues in Fig. 7. The 2-norm is often the `natural' norm to use in applications, but if the matrix **A** is large, its computation is costly ($O(mn^2)$ floating point operations). For quick estimates that only require O (mn) operations and are accurate to a factor of $\sqrt{m}$ or $\sqrt{n}$, use the matrix 1- or $\infty$-norms. The SVD has many other important uses.

For example, if $\mathbf{A} \in \mathbf{C}^{n\times m}$ is invertible and non singular, we have $\mathbf{A}^{-1} = \mathbf{V}\mathbf{\Sigma}^{-1}\mathbf{U}^{\mathbf{T}}$, and so

$$\|\mathbf{A}^{-1}\|_2 = \frac{1}{\min\limits_{\|x\|_2=1}\|\mathbf{A}x\|_2} = \frac{1}{\sqrt{\lambda_{\min}}} = \frac{1}{\sigma_n} \tag{3}$$

where $\lambda_{\min}$ is the smallest eigenvalue such that $A^T A - \lambda I$ is singular. This illustrates that a square matrix is singular if and only if $\sigma_n = 0$.

We shall explore this in more details later using the SVD to construct low-rank approximations to $A$.

## 3. LOW-RANK MATRIX APPOXIMATION

For simplicity, assume $m \geq n$. Then the SVD of A can be written as $A = U\Sigma V^T$ can be written as the linear combination of m-by-n outer product matrices:

$$\mathbf{A} = \mathbf{U}\Sigma\mathbf{V}^{\mathbf{T}} = \begin{bmatrix}\sigma_1 u_1 & \sigma_2 u_2 & \dots & \sigma_n u_n\end{bmatrix}\begin{bmatrix} v_1^T \\ v_2^T \\ \vdots \\ v_n^T \end{bmatrix} = \sum_{j=1}^{n} \sigma_j u_j v_j^T \tag{4}$$

One of the key applications of the singular value decomposition is the construction of low-rank approximations to a matrix. Hence for any $x \in \mathbf{C}^n$

$$\mathbf{A}x = \sum_{j=1}^{n} (\sigma_j u_j v_j^T) x = \sum_{j=1}^{n} (\sigma_j v_j^T x) u_j \tag{5}$$

since $v_j^T x$ is just a scalar. We see that $\mathbf{A}x$ is linear combination of the left singular vectors $\{\mathbf{u}_j\}$. The only catch is that $u_j$ will not contribute to the above linear combination if $\sigma_j = 0$. If all the singular values are nonzero, set r = n; otherwise, define r such that $\sigma_r \neq 0$ but $\sigma_{r+1} = 0$. Then we have low-rank approximation to a matrix **A**:

$$\mathbf{A}\mathrm{x} = \sum_{j=1}^{r} (\sigma_j \mathrm{v}_j^T \mathrm{x}) \mathrm{u}_j \tag{6}$$

We can approximate A by taking only a partial sum here:

$$\mathbf{A_k} = \sum_{j=1}^{k} \sigma_j \mathrm{u}_j \mathrm{v}_j^T \quad \text{or} \tag{7a}$$

$$\mathbf{A_k} = \sum_{j=1}^{k} \sigma_j \mathbf{A_j} \tag{7b}$$

for $k \leq r$. The linear independence of $\{\mathrm{u}_1, \cdots \mathrm{u}_k\}$ guarantees that rank($\mathbf{A}_k$) = k with an $m \times n$ a weight matrix $\mathbf{A_j^w} = \mathrm{u}_j \mathrm{v}_j^T$. Example, we can see that for $\mathrm{j} = 5$, where eigenvalues are:
$\sigma_1 = 794.3, \sigma_2 = 71.841,\ \sigma_3 = 30338,\ \sigma_4 = 29.161,\ \sigma_5 = 16.107$.

Therefore we obtained matrix decomposition:

$$\mathbf{A_5} = \sigma_1 \mathrm{u}_1 \mathrm{v}_1^T + \cdots + \sigma_5 \mathrm{u}_5 \mathrm{v}_5^T = \sigma_1 \mathbf{A_1^w} + \sigma_2 \mathbf{A_2^w} + \cdots + \sigma_5 \mathbf{A_5^w} \tag{8}$$

and we can represent image with weight matrices (Fig.2) and example of (Fig.3) for j=5. Thus the terms $\sigma_j \mathrm{u}_j \mathrm{v}_j^T$ with small $\sigma_j$ contribute very little to original matrix. We can get rid of them and still to have a good approximation to matrix $\mathbf{A}$.

But how well does this partial sum approximate A? This question is answered by the following result (Nievergelt, 1997) that has wide-ranging consequences in applications.

**Theorem:** For all $1 \leq \mathrm{k} < \mathrm{rank(A)}$,

$$\min_{\mathrm{rank(X)=k}} \|\mathbf{A} - \mathbf{X}\| = \sigma_{k+1} \tag{9}$$

with the minimum attained by $\mathbf{A_k} = \sum_{j=1}^{k} \sigma_j \mathrm{u}_j \mathrm{v}_j^T$

There is a full proof description in Yang and Lu (1995). Notice that we do not claim that the best rank-k approximation given in the theorem is unique.

## 4. EXAMPLE OF THE SINGULAR VALUE DECOMPOSITION

The standard algorithm for computing the singular value decomposition differs a bit from the algorithm described above. We know from our experiences with the normal equations for least squares problems that significant errors can be introduced when $A^T A$ are constructed. For practical SVD computations, one can sidestep this by using Householder transformations to create unitary matrices U and V such that $B=UAV^T$ is bidiagonal, i.e., $b_{jk} = 0$ unless $j = k$ or $j-1=k$. One then applies specialized eigenvalue algorithms for computing the SVD of a bidiagonal matrix. While this approach has numerical advantages over the method used in our constructive proof of the SVD, it is still instructive to follow through that construction for a simple matrix, say

$$\mathbf{A} = \begin{bmatrix} 0 & 1 \\ 1 & 0 \\ 1 & 1 \end{bmatrix}$$

**Step 1.** First, form $\mathbf{A}^\mathbf{T}\mathbf{A}$ :

$$\mathbf{A}^\mathbf{T}\mathbf{A} = \begin{bmatrix} 2 & 1 \\ 1 & 2 \end{bmatrix}$$

and compute its eigenvalues, $\lambda$ , and (normalized) eigenvectors, $\mathrm{v}$ :

$$\lambda_1 = 3\,, \mathrm{v}_1 = \frac{1}{\sqrt{2}}\begin{bmatrix} 1 \\ 1 \end{bmatrix} \text{ and } \lambda_2 = 1\,, \mathrm{v}_2 = \frac{1}{\sqrt{2}}\begin{bmatrix} 1 \\ -1 \end{bmatrix}$$

**Step 2**. Set

$$\sigma_1 = \left\|\mathbf{A}v_1\right\|_2 = \sqrt{\lambda_1} = \sqrt{3} \text{ and } \sigma_2 = \left\|\mathbf{A}v_2\right\|_2 = \sqrt{\lambda_2} = 1$$

**Step 3**. Since $\sigma_1, \sigma_2 \neq 0$ , we can immediately form $\mathrm{u}_1$ and $\mathrm{u}_2$

$$\mathrm{u}_1 = \frac{1}{\sigma_1}\mathbf{A}\mathrm{v}_1 = \frac{1}{\sqrt{6}}\begin{bmatrix} 1 \\ 1 \\ 2 \end{bmatrix} \quad \text{and} \quad \mathrm{u}_2 = \frac{1}{\sigma_2}\mathbf{A}\mathrm{v}_2 = \frac{1}{\sqrt{2}}\begin{bmatrix} -1 \\ 1 \\ 0 \end{bmatrix}$$

The $\frac{1}{\sigma_j}$ scaling ensures that both $\mathrm{u}_1$ and $\mathrm{u}_2$ are unit vectors. We can verify that they are orthogonal:

$$\mathrm{u}_1^{\mathrm{T}}\mathrm{u}_2 = \frac{1}{\sqrt{12}}\begin{bmatrix}1 & 1 & 2\end{bmatrix}\begin{bmatrix}-1\\ 1\\ 0\end{bmatrix} = 0$$

**Step 4.** At this point, we have all the ingredients to build the reduced singular value decomposition:

$$\mathbf{A} = \mathbf{U}\Sigma\mathbf{V}^{\mathbf{T}} = \begin{bmatrix}\frac{1}{\sqrt{6}} & -\frac{1}{\sqrt{2}}\\ \frac{1}{\sqrt{6}} & \frac{1}{\sqrt{2}}\\ \frac{2}{\sqrt{6}} & 0\end{bmatrix}\begin{bmatrix}\sqrt{3} & 0\\ 0 & 1\end{bmatrix}\begin{bmatrix}\frac{1}{\sqrt{2}} & \frac{1}{\sqrt{2}}\\ \frac{1}{\sqrt{2}} & -\frac{1}{\sqrt{2}}\end{bmatrix}$$

The only additional information required to build the full SVD is the unit vector $\mathrm{u}_3$ that is orthogonal to $\mathrm{u}_1$ and $\mathrm{u}_2$. One can and such a vector by inspection:

$$\mathrm{u}_3 = \frac{1}{\sqrt{3}}\begin{bmatrix}1\\ 1\\ -1\end{bmatrix}$$

If you are naturally able to eyeball this orthogonal vector, there are any number of mechanical ways to compute $\mathrm{u}_3$, e.g., by finding a vector $u_3 = [\alpha, \beta, \gamma]^T$ that satisfies:

Orthogonality conditions $\mathrm{u}_1^{\mathrm{T}}\mathrm{u}_3 = \mathrm{u}_2^{\mathrm{T}}\mathrm{u}_3 = 0$

Normalization condition $\mathrm{u}_3^{\mathrm{T}}\mathrm{u}_3 = 1$

We can find vector $u_3$ using the Gram-Schmidt process too.

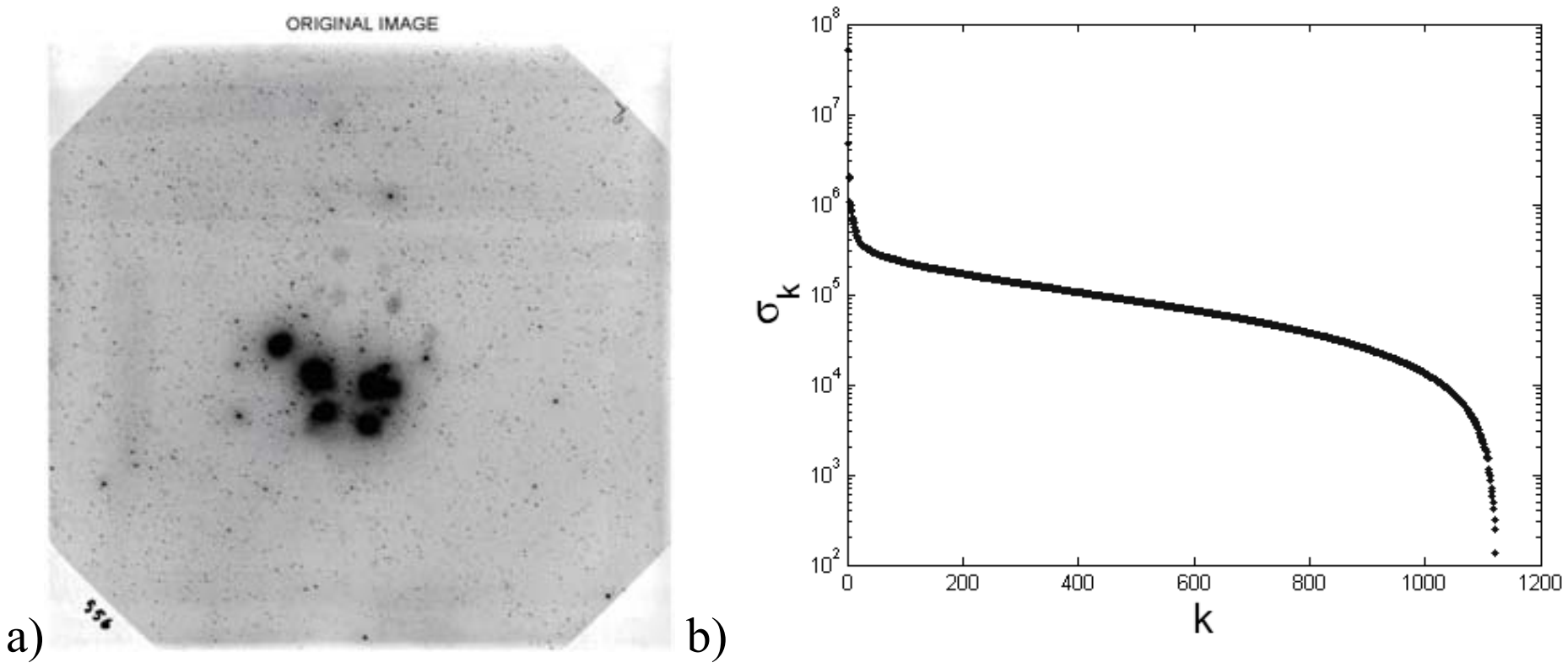

**Figure 4.** Image of SPP ASI067 000556 (M45-556p.fits) in the region of the Pleiades stellar cluster. a) Original image of SPP (size 1122x1122), b) Singular values.

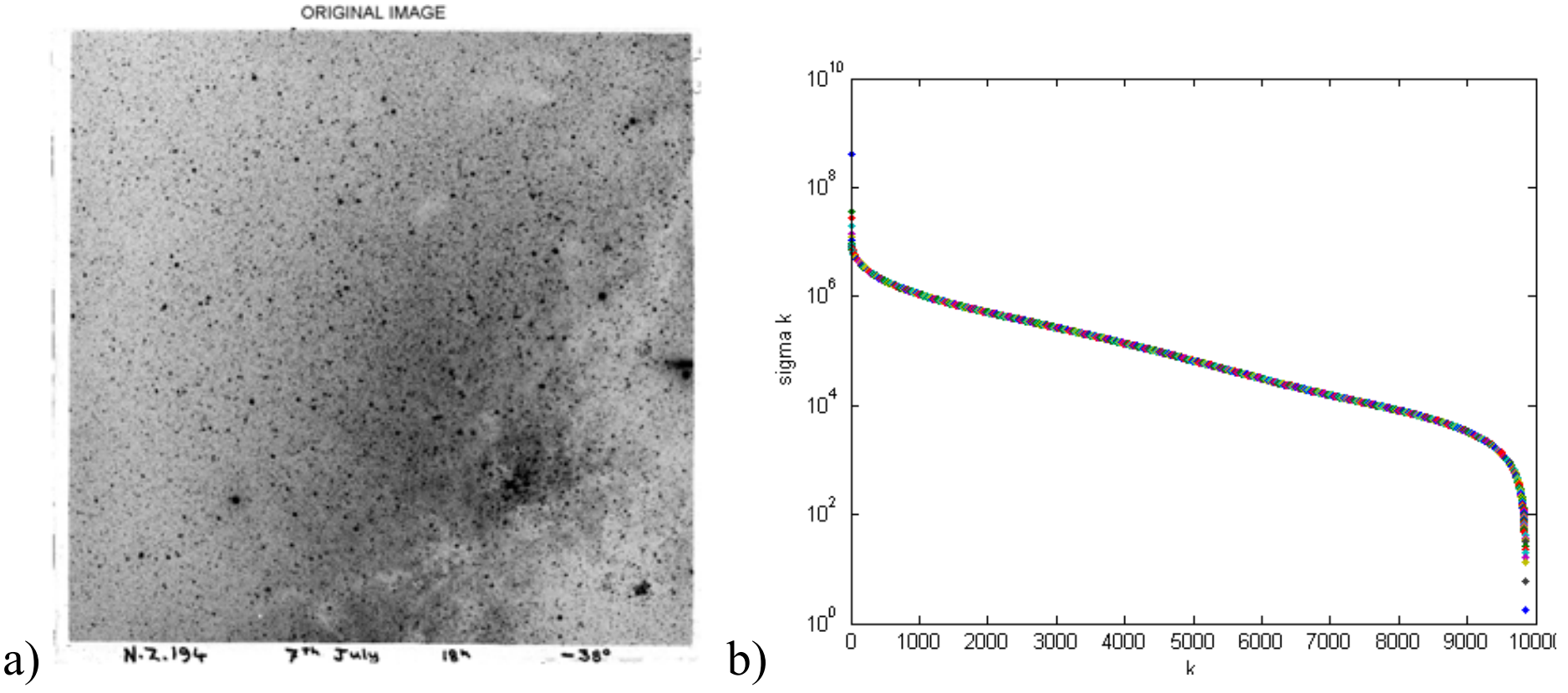

**Figure 5.** Image of SPP BAM010M (nz194.fits); a) Original image SPP (size 9898x9897); b) Singular values.

## 5. APPLICATION OF LOW-RANK APPROXIMATION IN IMAGE COMPRESION

As an illustration of the utility of low-rank matrix approximations, we consider the compression of digital images. On a computer, the image is simply a matrix denoting pixel colors. Typically, such matrices can be well approximated by low-rank matrices. Instead of storing the mn entries of the matrix A, one need only store the $k(m+n)+k$ numbers that make up the various $\sigma_j$, $u_j$ and $v_j$ values in the sum:

$$\mathbf{A_k} = \sum_{j=1}^{k} \sigma_j \mathrm{u}_j \mathrm{v}_\mathrm{j}^\mathrm{T} \tag{10}$$

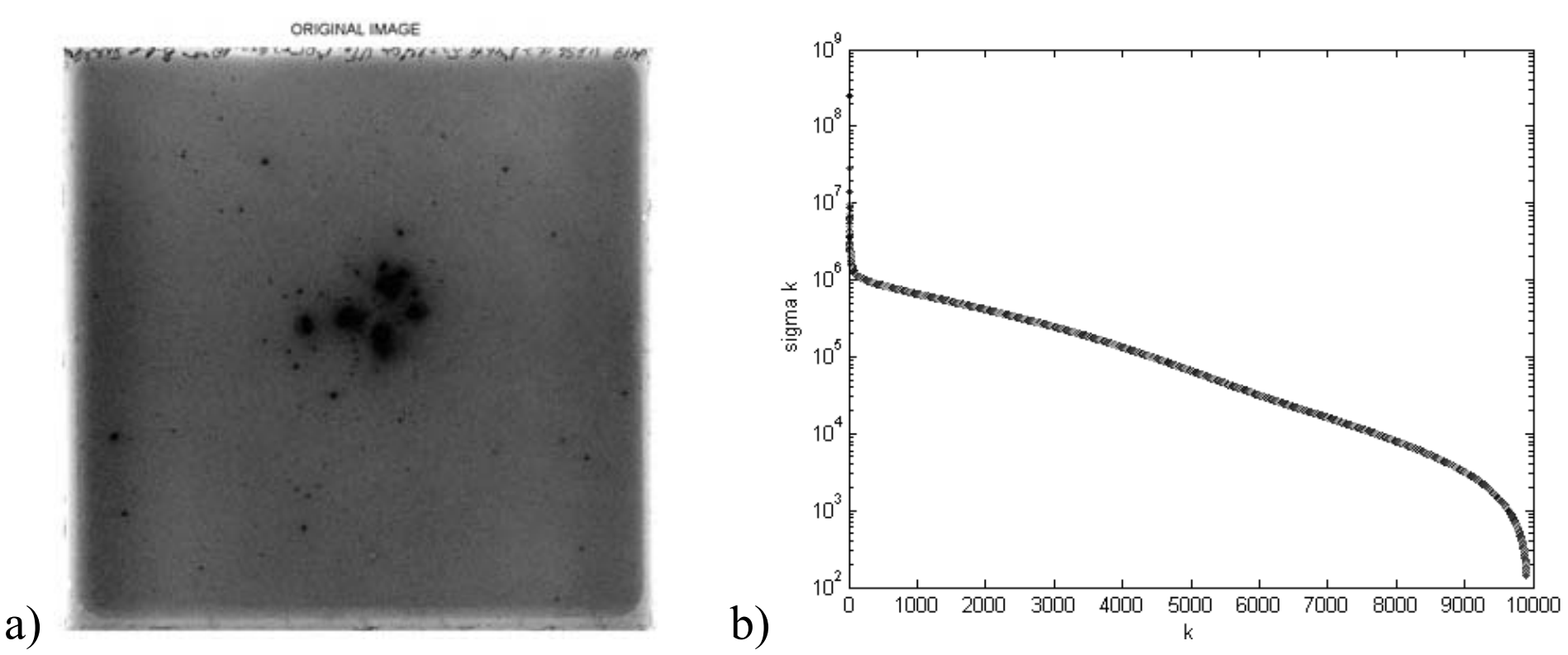

**Figure 6.** Image of SPP ROZ050 006419 (6419.fits) in the region of the Pleiades stellar cluster a) Original image SPP (size 9906x10060); b) Singular values.

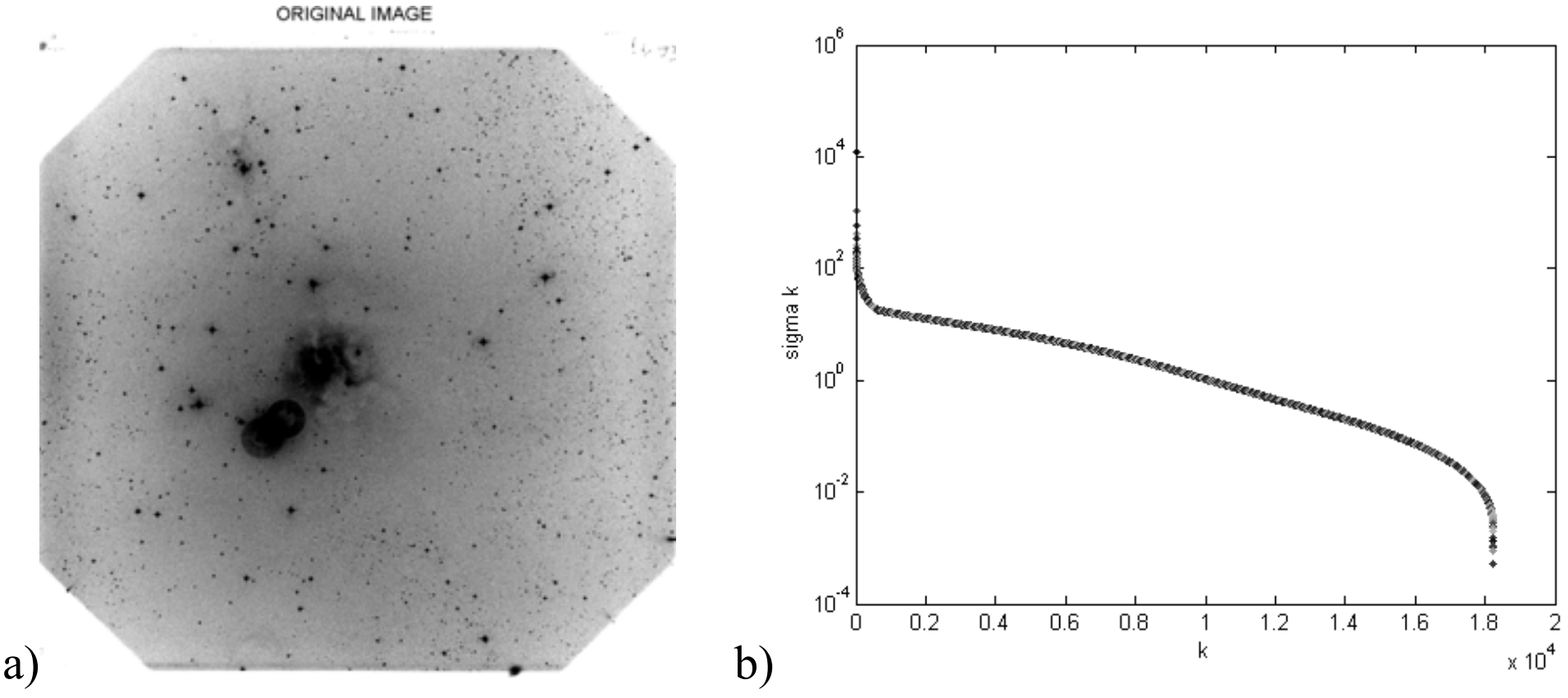

**Figure 7.** Image of SPP ROZ200 001655 (ROZ200 001655a.fits), taken in the region of S MON a) Original image SPP (size 18898x18240); b) Singular values

When k<<min(m, n) this can make for a significant improvement, though modern image compression protocols use more sophisticated approaches.

Next, we show the singular values for one image matrix, the scanned plate ASI067 000556 (M45-556p.fits) in the region of the Pleiades stellar cluster, with image size 1112x1122 and 16bit pixel format. We see that the singular values decrease near linearly. Though the singular values are very large, $\sigma_1 > 10^7$, fig.4b, there is a relative difference of five orders of magnitude between the smallest and largest singular value. We see that the singular values decrease rapidly.

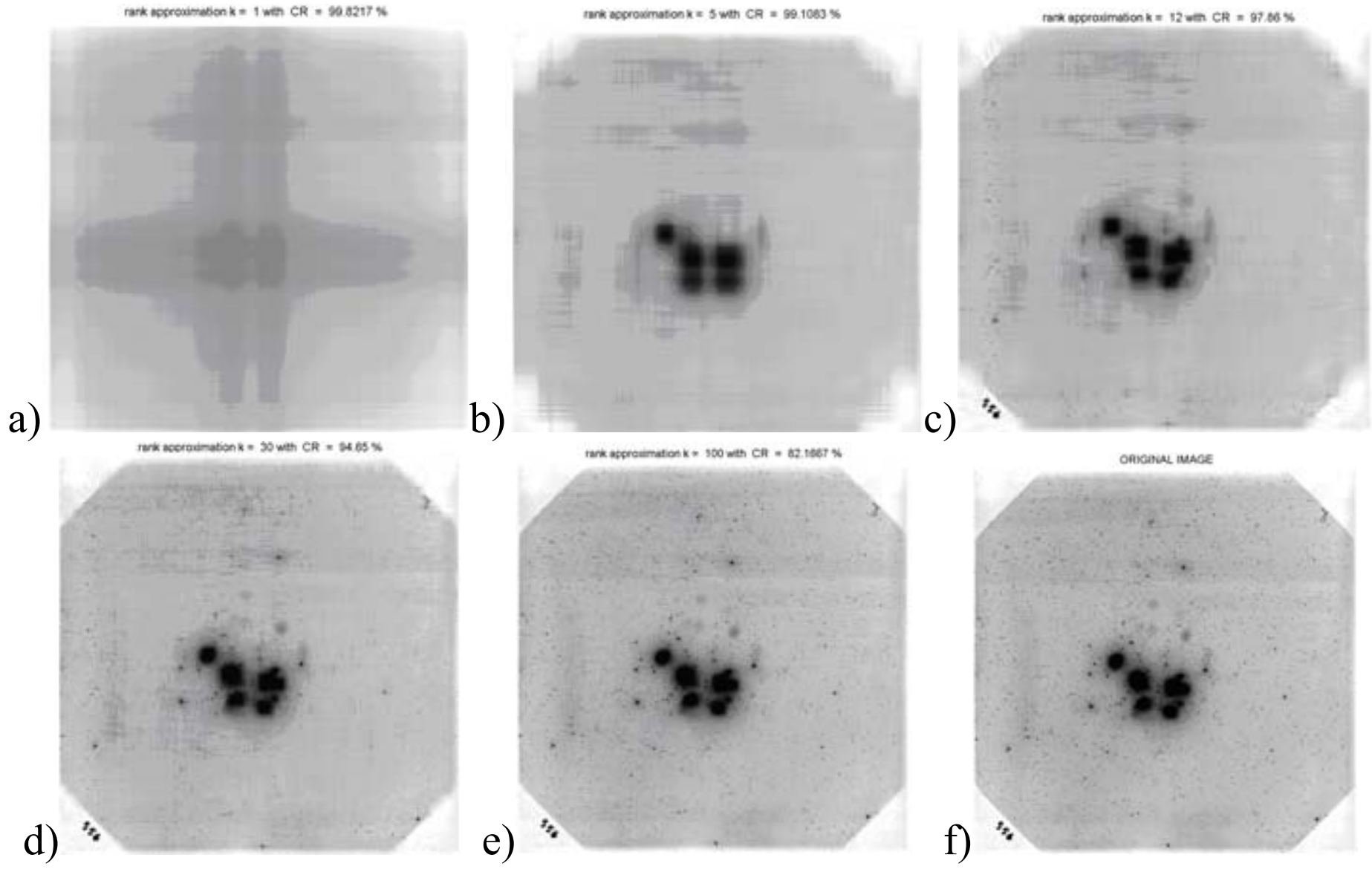

**Figure 8.** The SVD rank approximations for image SPP ASI067 000556 (M45-556p.fits) in the region of the Pleiades stellar cluster.

There are one greater than $10^7$ and only four greater than $10^6$. If all the singular values were roughly the same, we would not expect accurate low-rank approximations. We can approximate a matrix by adding only the first few terms of the series (Fig. 8).

For image quality measure we use compressed ratio. It is given with:

$$\mathrm{CR} \equiv \text{compression ration} = \left(1 - \frac{\mathrm{rank(A)}(\mathrm{sizeX(A)} + \mathrm{sizeY(A)} + 1)}{\mathrm{sizeX(A)}\,\mathrm{sizeY(A)}}\right) \times 100,\ \%$$

Thus, one way of compressing the image is to compute the singular value decomposition and then to reconstruct the image by an approximation of smaller rank.

This technique is illustrated in Fig. 8, which shows respectively the terms $\mathrm{u}_j \mathrm{v}_\mathrm{j}^\mathrm{T}$ and the terms $\mathbf{A_k} = \sum_{j=1}^{k} \sigma_j \mathrm{u}_j \mathrm{v}_\mathrm{j}^\mathrm{T}$. As can be seen in Fig. 8 and Fig. 9, the image is reconstructed almost perfectly (according to the human eye) by a rank 40 approximation. This gives a compression ratio of:

$$\mathrm{CR} = \left(1 - \frac{40(1122 + 1122 + 1)}{1122 \times 1122}\right) \times 100 = 0.928667 \times 100 \approx 93\,\%$$

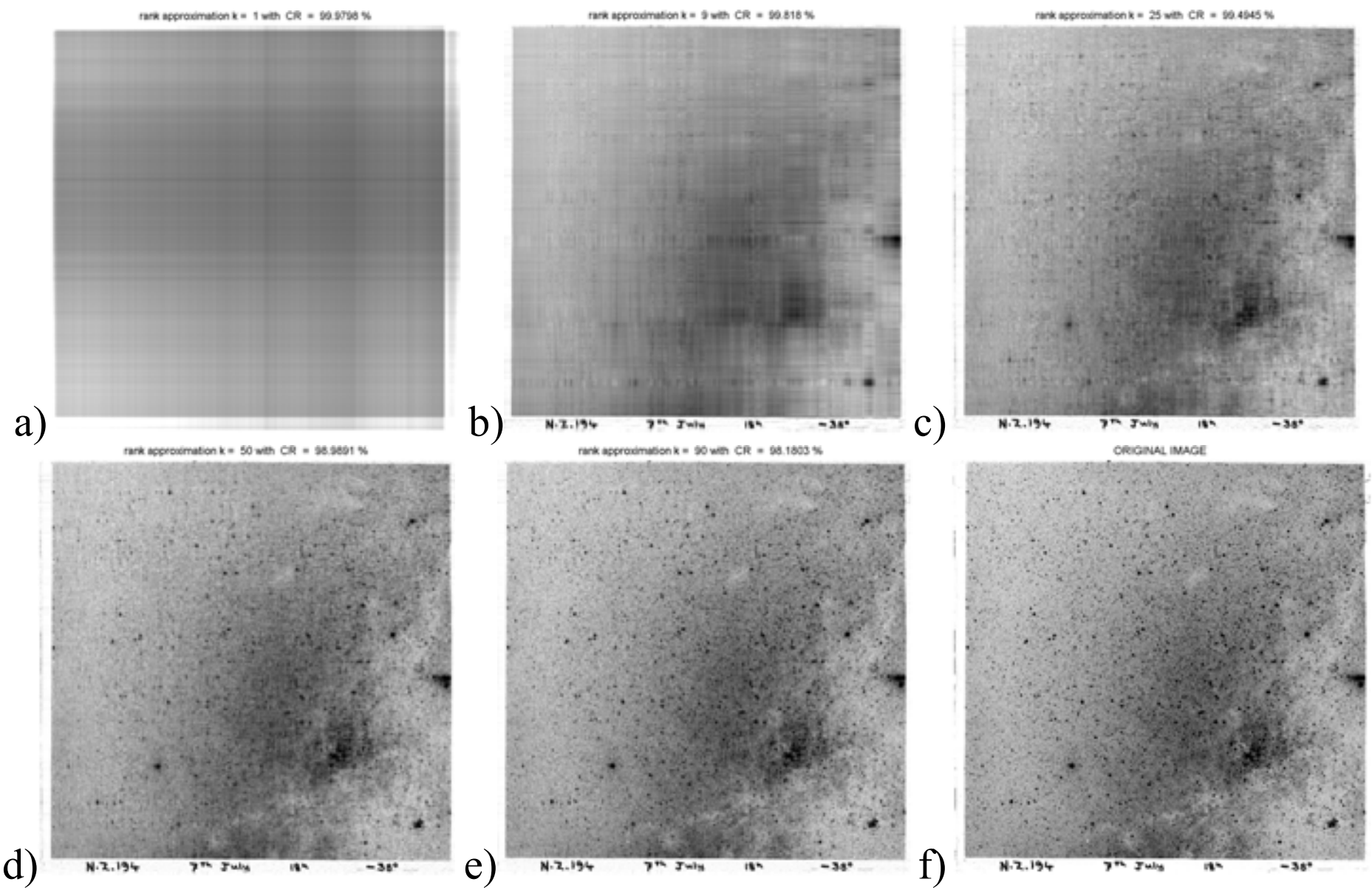

**Figure 9.** The SVD rank approximations for scanned image of SPP BAM010M (nz194.fits).

Let us to show rank approximation image matrix of a scanned plate BAM010M (nz194.fits), with image size 9898x9897 and 16bit pixel format. The first ten singular values are:

$\sigma_1 = 399\,935\,695$, $\sigma_2 = 36\,103\,983$, $\sigma_3 = 27\ \ 223\ \ 347$, $\sigma_4 = 19834987$,

$\sigma_5 = 13977320$ $\sigma_6 = 12295017$, $\sigma_7 = 10881892$, $\sigma_8 = 10418273$,

$\sigma_9 = 9\,556\,364$, $\sigma_{10} = 9037119$

…

$\sigma_{9849} = 5.954$, $\sigma_{9850} = 1.832$, and $\sigma_{9851} = 4.7 * 10^{-11}$!

We obtained that after a 9850 singular value all singular values are zeros (Fig. 5b).

Let us to show rank approximation image matrix of a scanned plate ROZ200 001655a.fits, with image size 18898x18240 and 16bit pixel format.

The first ten singular values are:

$\sigma_1 = 12246.0868$, $\sigma_2 = 1060.9436$, $\sigma_3 = 578.4546$, $\sigma_4 = 413.6548$,

$\sigma_5 = 333.3525$ $\sigma_6 = 267.0412$, $\sigma_7 = 222.5709$, $\sigma_8 = 199.2704$,

$\sigma_9 = 187.1444$, $\sigma_{10} = 183.6114$

… and $\sigma_{18240} = 5 * 10^{-4}$. The singular values are represented in Fig. 7b.

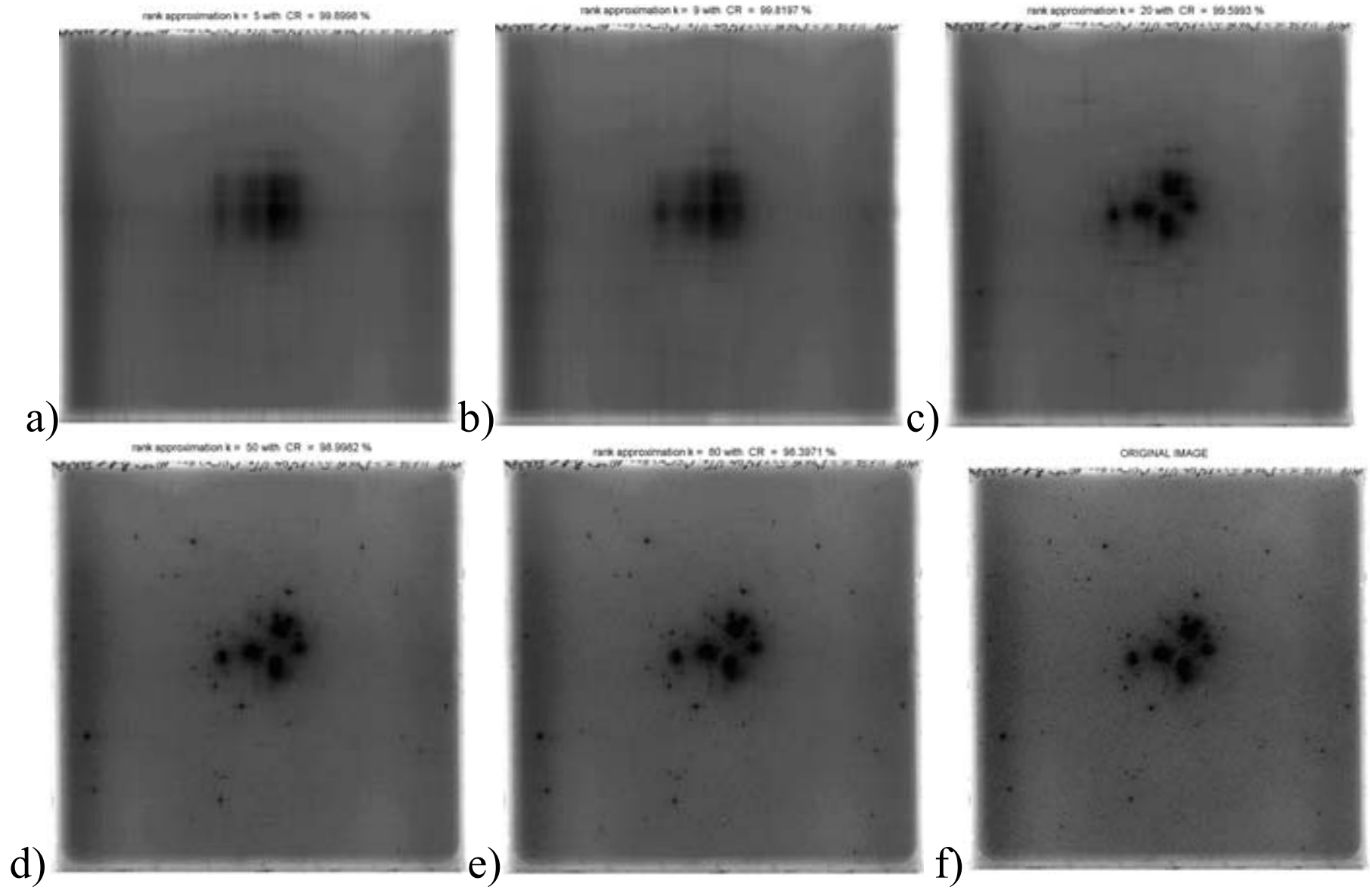

**Figure 10.** The SVD rank approximations for the scanned image of SPP ROZ050 006419 (6419.fits) taken in the region of the Pleiades stellar cluster.

## 6. CONCLUSIONS

Since matlab code for SVD calculate full rank SVD, we create matlab code for step by step rank approximation simulation for image processing of SPP.

As rank $k$ increases, the images quality increases, but the same does the volume of memory needed to store the images. This means that smaller ranked SVD approximations (smaller values for $j$) are preferable.

By storing only the first columns of $\mathbf{U}$ and $\mathbf{V}$ and their respective singular values, the image can be replicated while taking up only:

- 5.35% for image of CR=94.65% with k=30, image size (1122x1122) (Fig. 8d),
- 1.01% for image of CR=98.99% with k=50, image size (9898x9897) (Fig. 9d),
- 1% for image of CR=99.00% with k=50, image size (9906x10060) (Fig. 10d)

of the original storage space.

We can actually see how the compression breaks down the matrix for a rank approximation. Notice that every row of pixels is the same row, just multiplied by a different constant, which changes the overall intensity of row. The same goes for columns: every column of pixels is actually the same column multiplied by a different constant (Fig. 3).

Notable those with minimum number rank,

- rank 12 with CR=97.86, image size (1122x1122) (Fig. 8c),
- rank 9 with CR=99.82%, image size (9898x9897) (Fig. 9b),
- rank 9 with CR=98.82%, image size (9906x10060) (Fig. 10b),

we read clearly the marks of the plates. With very big CR we can see image details. This can be use similar as filter and useful for image denoising. The rank image approximation is faster from wiener filter processing. This is important when there are large images (e.g. scanned plates). These low-rank matrix approximations to SPP images do require less computer storage and transmission time than the full-rank image.

The SVD facilitates the robust solution of a variety of approximation problems, including not only the least squares problems with rank-deficient A, but also other low-rank matrix approximation problems that arise throughout engineering, statistics, the physical sciences, and social sciences too.

## Acknowledgements

The authors are grateful to Prof. O. Kounchev from the Institute of Mathematics and Informatics, Bulgarian Academy of Sciences for the advices and to A. Borisova for providing images of scanned plates in the Pleiades field.

This work has been supported by the research project DO-02-275 of the Bulgarian National Science Fund.